\documentclass[10pt,conference,final]{IEEEtran}
\usepackage{amsmath,graphicx,subfigure,amssymb,flushend}
\usepackage{algorithm,algorithmic,color,colortbl}
\usepackage{booktabs}
\title{Tumour Ellipsification in Ultrasound Images \\ for Treatment Prediction in Breast Cancer}

\author{Mehrdad~J.~Gangeh,~\IEEEmembership{Senior Member,~IEEE,}
        Hamid R. Tizhoosh, \\
        Kan Wu,
        Dun Huang,
        Hadi~Tadayyon,
        and Gregory~J.~Czarnota
\thanks{This work was supported in part by the Natural Science and Engineering Research Council (NSERC) of Canada under Postdoctoral Fellowship (PDF-454649-2014), in part by the Terry Fox Foundation, and in part by the Canadian Institutes of Health Research.}
\thanks{M.~J.~Gangeh, H.~Tadayyon, and G.~J.~Czarnota are with the Departments of Medical Biophysics, and Radiation Oncology, University of Toronto, Toronto, ON, M5G 2M9 Canada; also with the Departments of Radiation Oncology, and Imaging Research - Physical Sciences, Sunnybrook Health Sciences Centre, Toronto, ON, M4N 3M5 Canada (e-mails: mehrdad.gangeh@utoronto.ca, \{hadi.tadayyon,gregory.czarnota\}@sunnybrook.ca).}% <-this % stops a space
\thanks{H.~R.~Tizhoosh, K.~Wu, and D.~Huang are with KIMIA Lab, University of Waterloo, Waterloo, ON, N2L 3G1 Canada. (e-mails: \{tizhoosh,k8wu,dun.huang\}@uwaterloo.ca).}
}

\flushend
\begin{document}
\maketitle

\begin{abstract}
Recent advances in using quantitative ultrasound (QUS) methods have provided a promising framework to non-invasively and inexpensively monitor or predict the effectiveness of therapeutic cancer responses. One of the earliest steps in using QUS methods is contouring a region of interest (ROI) inside the tumour in ultrasound B-mode images. While manual segmentation is a very time-consuming and tedious task for human experts, auto-contouring is also an extremely difficult task for computers due to the poor quality of ultrasound B-mode images. However, for the purpose of cancer response prediction, a rough boundary of the tumour as an ROI is only needed. In this research, a semi-automated tumour localization approach is proposed for ROI estimation in ultrasound B-mode images acquired from patients with locally advanced breast cancer (LABC). The proposed approach comprised several modules, including 1) feature extraction using keypoint descriptors, 2) augmenting the feature descriptors with the distance of the keypoints to the user-input pixel as the centre of the tumour, 3) supervised learning using a support vector machine (SVM) to classify keypoints as ``tumour'' or ``non-tumour'', and 4) computation of an ellipse as an outline of the ROI representing the tumour. Experiments with 33 B-mode images from 10 LABC patients yielded promising results with an accuracy of 76.7\% based on the Dice coefficient performance measure. The results demonstrated that the proposed method can potentially be used as the first stage in a computer-assisted cancer response prediction system for semi-automated contouring of breast tumours.   % which outperformed some rival methods including \emph{k}-means and fuzzy \emph{c}-means.
\end{abstract}

%\begin{keywords}
%Breast cancer, response prediction, ultrasound, segmentation, feature descriptors, keypoints.
%\end{keywords}

\section{Idea and Motivation}\label{sec:intro}

Precision medicine is an emerging technique that tailors medical treatment depending on the characteristics of each patient~\cite{PM:Collins15}. For example, in cancer therapy, the main goal of precision medicine is to decide on the most effective cancer therapy for a patient based on his/her response to treatment administrations. To this end, the first step in personalized cancer therapy is to either predict or provide a means to assess the individual responses to treatment early during the course of a therapy. There are several functional imaging modalities such as magnetic resonance imaging (MRI), diffuse optical spectroscopy (DOS), and positron emission tomography (PET) that can provide imaging at a microscopic level to detect cell death~\cite{MI:Brindle08}. Two main limitations of these imaging technologies include: the requirements for a large capital investment and an external agent. The latter is also expensive, and may cause some side effects and/or allergic reactions~\cite{MI:Oelze16}. In contrast, quantitative ultrasound (QUS) methods provide a portable, non-expensive, and non-invasive means for a rapid acquisition of functional images that can be used for an early assessment of cancer therapeutic effects~\cite{MI:Oelze16}. Moreover, in QUS methods, the endogenous contrast -- generated by the very process of cell death -- is employed in treatment assessment, which alleviates the requirement for injecting external agents.

The applications of QUS methods have recently been extended from cancer response monitoring \cite{Gangeh:TMI16} to cancer response prediction \cite{MI:Tadayyon15}, and tissue characterization/visualization using 3-D automated breast ultrasound (ABUS) scanners \cite{Gangeh:SPIE16}. The first major step in the implementation of each of these applications is to contour a region of interest (ROI) inside the tumour in frames with identifiable tumour areas. This step is currently performed manually as there is no automated software to segment an ROI in ultrasound B-mode images. The manual segmentation of tumours, however, is a very time-consuming task as hundreds of frames should be contoured in a study due to the availability of multiple frames for each subject (an animal in preclinical and a patient in clinical studies). With the availability of 3-D scanners such as ABUS technologies, the problem will be even more severe as tens of frames should be contoured in each patient. Therefore, designing an automated segmentation method can save a significant amount of experts' time and efforts.

In this study, a semi-automated supervised tumour localization method was proposed for ROI estimation in B-mode images acquired from patients with locally advanced breast cancer (LABC).% \cite{LABC:Newman09}.

%The paper is organized as follows: Section \ref{sec:bkg} briefly reviews the relevant methods. The proposed approach is described in section \ref{sec:prop}. Experiments and results are reported in section \ref{sec:exp}.

%---------------Intro by students
\section{Background}
\label{sec:bkg}

There is an extensive literature on image segmentation. Several segmentation techniques have been tried for ultrasound images~\cite{misc:Sridevi}. In a more recent literature, classification techniques applied on some expressive features have been more frequently employed to extract segments from images. These classification techniques include simple methods like \emph{k}-means and fuzzy \emph{c}-means (FCM) and more sophisticated algorithms such as support vector machines (SVM). %~\cite{misc:Cortes95}. 
These classifiers group pixels into segments to enable subsequent measurements and recognition. Most of these methods, when applied to digital images, rely on the availability of invariant features. Among feature extraction methods in the literature, those which are based on invariant features extracted from some \emph{keypoints} demonstrated success in segmentation tasks. Keypoints represent pixel positions in an image surrounded (within certain vicinity) with significant information, e.g., textures, corners, edges, etc~\cite{misc:lowe04}. For instance, FAST (features from accelerated segment test)~\cite{misc:Rosten05}, BRISK (binary robust invariant scalable keypoints)~\cite{misc:Leutenegger11}, and SURF (speeded up robust features)~\cite{misc:Bay08} are among the most commonly used keypoint-based feature descriptors providing inputs to a classifier.

SURF is a feature detection method, which is partly inspired by the scale-invariant feature transform (SIFT) descriptor, but is generally faster than SIFT~\cite{misc:Bay08}. It uses an integer approximation of the determinant of the Hessian blob detector to detect keypoints. Its feature descriptor is based on the sum of the Haar wavelet response around the point of interest. The three  main modules of the algorithm include: interest point detection, local neighborhood descriptor, and matching.

In BRISK detector, the scale of each keypoint is estimated in the continuous scale-space~\cite{misc:Leutenegger11}. The BRISK descriptors are composed of binary vectors by concatenating the results of brightness comparisons. The key concept of the BRISK descriptor is based on using a specific pattern to sample the neighborhood of all keypoints.

FAST is a corner detection method~\cite{misc:Rosten05}. The advantage of using FAST corner detector is its high computational efficiency and a better performance when subsequent machine-learning methods such as a supervised classification are applied~\cite{misc:Rosten10}. It uses a 16-pixel circle to classify a candidate point as a corner or no corner, where for each pixel in the circle, an integer number from 1 to 16 will be labeled clockwise.

\section{Proposed Approach}
\label{sec:prop}
%In treatment prediction, one can predict the response of the tumour in QUS parametric maps by investigating multiple images available from each patient~\cite{MI:Tadayyon15}. 
The first step in tumour response prediction is contouring (segmenting) the tumours.  Subsequently, texture analysis will be performed on the QUS parametric maps computed within the contoured tumours (tumour cores), and their margins (including surrounding healthy tissue) to predict patient's response~\cite{MI:Tadayyon15}. In the aforementioned analysis steps, however, a rough localization of the tumour is of more interest than its exact segmentation~\cite{MI:Tadayyon15}. Considering this, and also the fact that ultrasound images are known to be very difficult for segmentation, a semi-automated approach was proposed in this study to obtain an estimation of the tumour location as described in the next paragraphs.

Algorithm \ref{alg:alg1} describes the proposed approach.

%% Proposed Approach -------------------------
%\begin{algorithm}[h!]
%\caption{Proposed Approach}
%\begin{algorithmic}[1]
%\label{alg:alg1}
%\STATE	-------- Processing --------
%\STATE	Read all images $I_i$
%\STATE	Apply hyperbolization to all images, $\beta=4$;
%\STATE	Apply median filter, mask size is $3\times 3$
%\STATE	Save images
%\STATE	-------- Training and classification --------
%\STATE	Extract BRISK/FAST/SURF features from $I_i$
%\STATE	Save locations of all keypoints
%\STATE	Save 64 descriptors of all keypoint $S_1,S_2,\dots$
%\STATE	Give the center points $C_i$ for all images
%\STATE	Calculate adjusted distance $D_i$ for each keypoint
%\STATE	List 64*$D_i$ as the 65th feature $S_65$
%\STATE	$N\leftarrow$ number of images
%\STATE	$S_k \leftarrow$ center point for $S_j$ using K-Means
%\STATE	$S_c\leftarrow$ center point for $S_j$ using Fuzzy C-Means
%\WHILE{$i<N$}
%	\STATE The features $S_j$ of images $I_i$ for testing.
%	\STATE The features $S_j$ of remaining training images
%	\STATE SVM training give the SVM structure
%	\STATE K-Means give $S_k$
%	\STATE Fuzzy C-Means give $S_c$
%	\STATE Use training result for testing
%	\STATE Save the classification results for image  $I_i$
%\ENDWHILE
%\STATE	--------  Accuracy measurement --------
%\WHILE{$i<N$}
%	\STATE	    Plot the ellipse $E_i$ for image $I_i$  based on the classification result
%	\STATE	    Compare $E_i$ and the actual tumour given by doctors
%	\STATE	    Using Dice/Jaccard for accuracy measurement
%\ENDWHILE
%\end{algorithmic}
%\end{algorithm}

% Proposed Approach -------------------------
\begin{algorithm}[h!]
\caption{Proposed Approach}
\begin{algorithmic}[1]
\label{alg:alg1}
\STATE	-------- Preprocessing and Feature Extraction --------
\STATE Read number of training images $N$
\STATE Calculate the average width $\bar{w}$ and height $\bar{h}$ of all segments
\WHILE{$i<N$}
	\STATE Read current image $\mathbf{I}_i$ and its ground truth image $\mathbf{G}_i$
	\STATE Pre-process $\mathbf{I}_i$ (contrast adjustment and filtering)
	\STATE Extract $n$ keypoints with $m$ descriptors $\mathbf{F}_i(n,m)$
	\STATE Acquire a point $(x_c^i,y_c^i)$ close to the centre of $\mathbf{G}_i$
	\WHILE {$j<n$}
		\STATE Calculate the weighted distance $d$ between  $(x_c^i,y_c^i)$ and the coordinates of keypoint $n_j$:
		\STATE $d_j = \sqrt{(x_j - x_c^i)^2 + [\frac{\bar{w}}{\bar{h}}(y_j - y_c^i)]^2}$
		
		\STATE Add distance to the descriptors $\mathbf{F}_i(n_j,m+1)=d_j$
	\ENDWHILE
	\STATE Save $\mathbf{F}_i(n,m'), (m'=m+1)$
\ENDWHILE
\STATE	--------  Testing --------
\WHILE {$i<N$}
	\STATE Read the current image $I_i$
	\STATE Use remaining images to train a classifier with features $\mathbf{F}(n,m')$ and target $E$
	\STATE Classify pixels of $I_i$ using the trained classifier
	\STATE Fit an ellipse $E_i$ into the tumour pixels $(x_1^i,y_1^i), (x_2^i,y_2^i), \dots$ provided by classifier
	\STATE Read  $\mathbf{G}_i$
	\STATE Calculate the accuracy $D_i=$Dice($\mathbf{G}_i, E_i$)
\ENDWHILE
\end{algorithmic}
\end{algorithm}

\textbf{Patient Data -- } The study involved 10 LABC patients with tumour sizes between 5 and 15~cm. The data acquisition was performed in accordance with the clinical research ethics approved by Sunnybrook Health Sciences Centre. A biopsy was used as the gold standard test to confirm all cancer cases. In order to measure the size of tumours, all patients were imaged using MRI. Also, all patients were imaged using ultrasound before the start of neoadjuvant chemotherapy (``pre-treatment''). The acquisition of US data was performed using a Sonix RP ultrasound system. The system was equipped with an L14-5/60 linear transducer (centre frequency at $\sim$7~MHz). Depending on tumour size and location, the transducer was focused at the midline of the tumour with a maximum depth of 4-6~cm. The number of scans obtained from each tumour depended on it size. This resulted in three to five scan planes from each tumour, with a scan plane separation of about 1~cm.

\textbf{Pre-Processing --} In order to increase the quality and contrast of B-mode images, before computation of feature descriptors, each image was pre-processed by applying the fuzzy histogram hyperbolization~\cite{Tizhoosh1995}. Subsequently, a 3 $\times$ 3 median filter was applied to each image.

\textbf{Feature Extraction --} Features were extracted from the keypoints identified on the pre-processed images for submission to a classifier. Three major feature extraction algorithms explained in Section~\ref{sec:bkg}, including SURF, FAST, and BRISK were adapted in this stage.

\textbf{Extended Features Descriptors --} After applying feature extraction methods, a feature descriptor of size $m$ for each keypoint was used to characterize an image. However, the preliminary results using these features revealed that there was no strong correlation between the tumour and the feature points to obtain an acceptable result (the initial performance using only extracted features resulted in relatively low performance, i.e., less than 60\% accuracy for segmentation of tumours). Although the manual contouring of multiple scans is a very time-consuming and tedious task, requiring the clinician to provide the rough location of the tumour centre (e.g., via a mouse click) does not pose an unacceptable burden on him/her. Assuming that the tumour centre  $(x_c,y_c)$ is available, the distance $d$ of each keypoint from $(x_c,y_c)$ can be added to the descriptor to augment the feature space (Algorithm~\ref{alg:alg1}, lines 9-12). This, as our experiments demonstrated later, considerably increased the accuracy of contouring tumours.

\textbf{Classification --} A support vector machine (SVM) classifier with a radial basis function (RBF) kernel was used in a supervised paradigm to classify the keypoints in each image as ``tumour'' or ``non-tumour'' points based on the extended feature descriptors $\mathbf{F}$ extracted from the keypoints (Algorithm~\ref{alg:alg1}, line ~14). In addition, \emph{k}-means and FCM were also used for the classification of the keypoints in the same way for the purpose of comparison with the SVM performance.

\textbf{Ellipsification --} After classification, all keypoints would be classified as ``tumour'' or ``non-tumour'' pixels. Suppose that there are $n$ points which are defined as tumour points in one image. For the purpose of ellipsification of a tumour area, at first, the positions of all the $n$ points were found in the vertical direction and the leftmost position $x_l$ and the rightmost position $x_r$ were identified. In the horizontal direction, the highest position $y_h$ and the lowest position $y_l$ were determined. These four points are the four vertices of an ellipse, which was used as the boundary of the ROI representing the tumour.

\textbf{Accuracy Measurement --} Dice coefficient was measured for calculating the accuracy of classification results. The Dice coefficient is given by
\begin{equation}
		D=\frac{2|E\cup G|}{|E|+|G| }	
\end{equation}		
where $E$ is the fitted ellipse and $G$ is the ground-truth. It can be interpreted as a measure of overlapping between the estimated and correct classes~\cite{Labatut2015arxiv}.

\textbf{Validation --} Leave-one-out validation was performed to evaluate the performance of the designed ellipsification system on the 33 available B-mode images. The algorithm was developed using Matlab (R2011a, Mathworks, USA) on a 64-bit Intel Core i5-4200U CPU @ 2.30~GHz processor equipped with 8~GB of memory and Windows 8.1.

\begin{figure}[tb]
\centering     %%% not \center
\includegraphics[width=2.8cm,height=2.8cm]{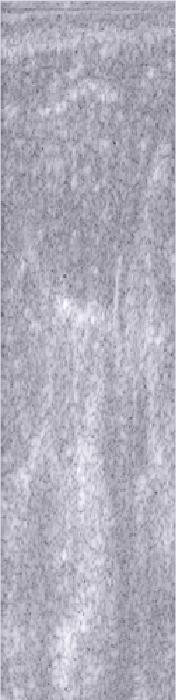}
\includegraphics[width=2.8cm,height=2.8cm]{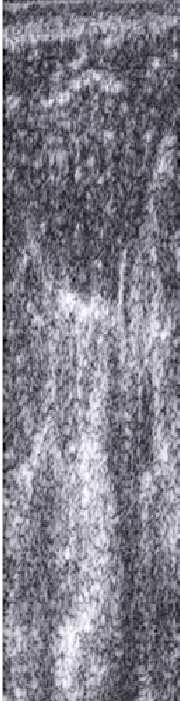}
\caption{A sample breast ultrasound image (left) and its enhanced pre-processed version after contrast adjustment using fuzzy histogram hyperbolization (right).}
\label{fig:hyperbol}
\end{figure}

\begin{figure}[tb]
\centering     %%% not \center
\includegraphics[width=1.8cm,height=1.8cm]{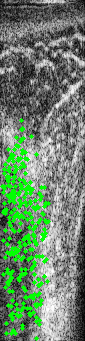}
\includegraphics[width=1.8cm,height=1.8cm]{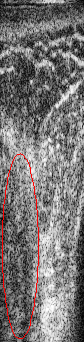}
\includegraphics[width=1.8cm,height=1.8cm]{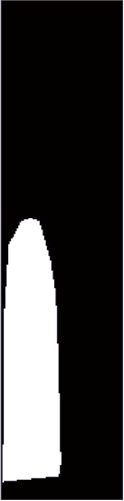}\\ \vspace{0.05in}
\includegraphics[width=1.8cm,height=1.8cm]{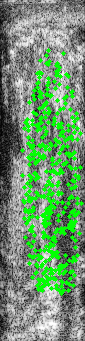}
\includegraphics[width=1.8cm,height=1.8cm]{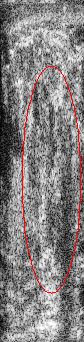}
\includegraphics[width=1.8cm,height=1.8cm]{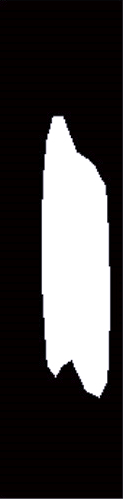}\\ \vspace{0.05in}
\includegraphics[width=1.8cm,height=1.8cm]{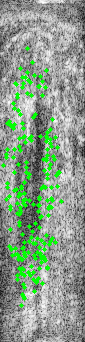}
\includegraphics[width=1.8cm,height=1.8cm]{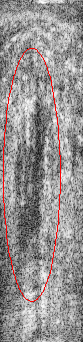}
\includegraphics[width=1.8cm,height=1.8cm]{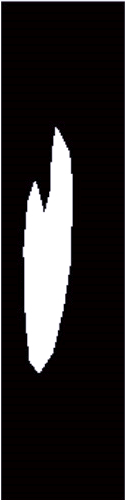}\\
\caption{Representative results for good, average, and bad ellipsification performances: 92.66\% (\emph{top}), 77.23\% (\emph{middle}), and 46.94\% (\emph{bottom}). Figure shows the computed keypoints (\emph{left}), estimated ellipses (\emph{middle}), and ground-truth contours (\emph{right}) on representative B-mode images.}
\label{fig:goodbad}
\end{figure}

\section{Results}

Fig.~\ref{fig:hyperbol} depicts the result of pre-processing on one typical breast ultrasound image using fuzzy histogram heprbolization. A comparison between the pre-processed (right) and the original (left) images shows the improvement in the quality of the B-mode image after pre-processing.

Fig.~\ref{fig:goodbad} shows representative good, average, and bad ellipsification results. As can be observed from this figure, the accuracy can significantly vary. Images with an irregular or small-size shape may not be classified correctly (accuracy as low as 46.9\% in Fig.~\ref{fig:goodbad}-bottom row). In contrast, for images with a normal size and shape, the accuracy can be quite high (e.g., 92.7\% in Fig.~\ref{fig:goodbad}-top row).

Table I provides the results of ellipsification using the Dice measure on all 33 B-mode images for various feature descriptors and classifiers used in the experiments. The highest average accuracy is 76.7\% with a standard deviation of 15.2\% using the SURF features and SVM classifier. In terms of computational cost, the SVM took longer time than other methods (approximately 10 minutes for each image, or 5 hours in total), while the other two classification methods were relatively fast. The results of classification can be improved by increasing the number of images as using only 33 scans in the experiments restricted the learning capability of the proposed ellipsification system.

\begin{figure}[t!]
\centering     %%% not \center
\includegraphics[width=0.78\linewidth]{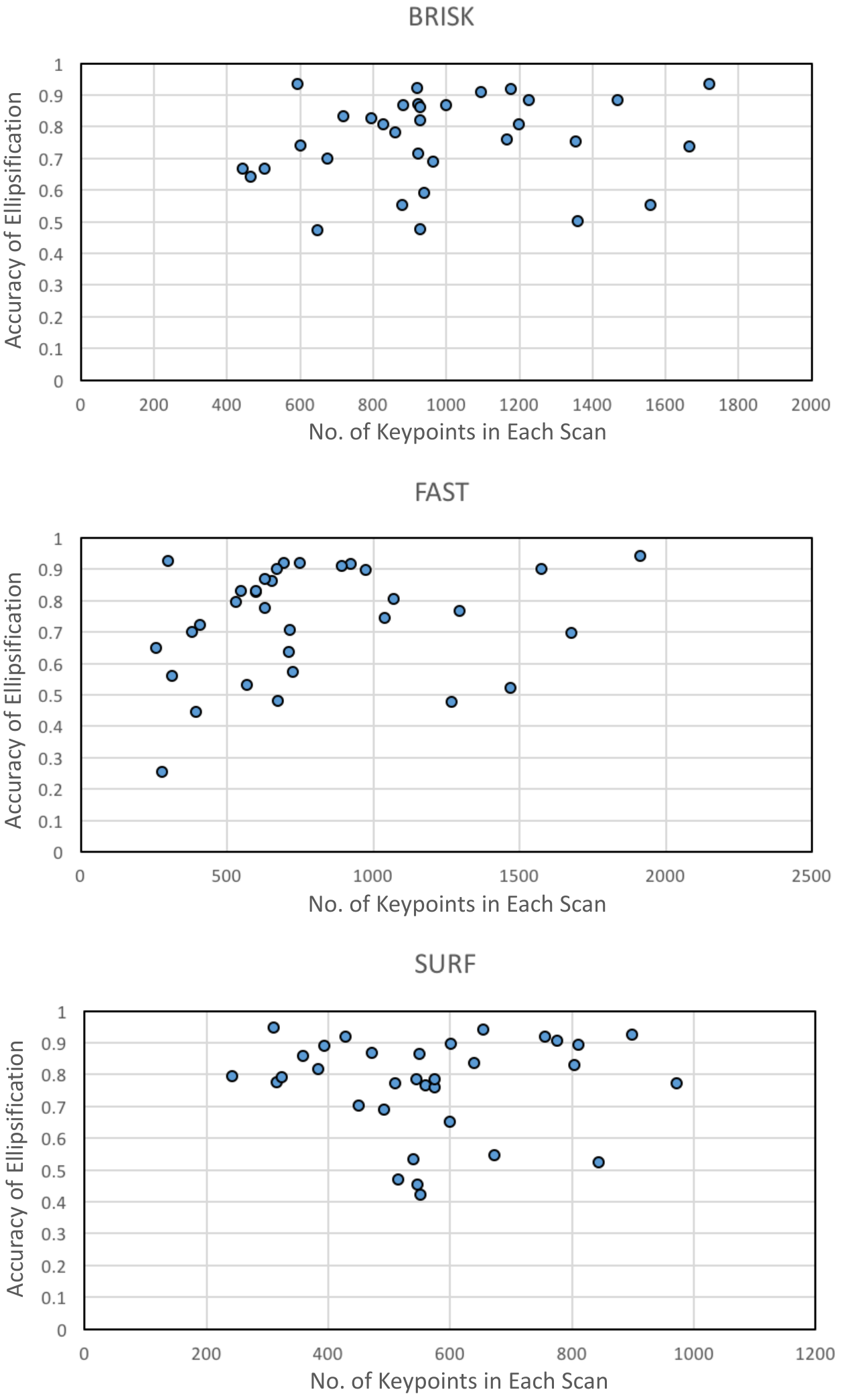}
\caption{The distribution of the accuracy of ellipsification versus the number of keypoints generated for all 33 B-mode images in the experiments. The slight advantage of SURF in our results (Table \ref{tab:results}) can be verified visually; compared to the BRISK and FAST, SURF appeared to result in higher accuracies more consistently, and within a narrower interval of features.}
\label{fig:accfeat}
\end{figure}

Fig. \ref{fig:accfeat} shows the distribution of the accuracy of ellipsification in respect to the number of generated keypoints for all the images used in the experiments (the results are only shown for the SVM classifier). BRISK, with an average accuracy of 75.63\%, produced $979\pm332$ keypoints. FAST, with an average accuracy of 73.66\%, resulted in $791\pm420$ keypoints. And SURF, with an average accuracy of 76.65\%, generated $565\pm179$ keypoints. Therefore, the SURF achieved a higher performance using a smaller number of keypoints, which reduced the computational cost of subsequent classification step. Moreover, the SURF was more consistent in selecting the number of keypoints from B-mode images, as can be judged from the smaller standard deviation (179) achieved compared to the other two methods.

%Fig. \ref{fig:goodbad} shows representative good, average, and bad ellipsification results. Apparently, the accuracy can significantly vary. Images with irregular shape or with a small size may not be classified correctly (accuracy as low as 46.94\%). In contrast, for images with a \emph{normal} size and shape, the accuracy can be quite high (e.g., 92.66\% as in Fig. \ref{fig:goodbad} ). This variability is most likely due to the variation in shape and size of the tumours. On the other hand, 33 cases are not sufficient to perform reliable classification. Relying on general knowledge in machine-learning domain, one may expect an increase in accuracy if more training data is available.

\begin{table}[]
\centering
\caption{The Results of ellipsification using the dice performance measure on 33 B-mode images based on ``leave-one-out'' validation scheme.}
\label{tab:results}
\begin{tabular}{llcl}
Features 		   & Classifier     & $D\pm \sigma$ 	\\ \hline
BRISK              & SVM            & 0.7563 $\pm$ 0.1387 \\
FAST               & SVM            & 0.7366 $\pm$ 0.1746  \\
SURF               & SVM            & 0.7665 $\pm$ 0.1515 \vspace{0.05in} \\ \hline
BRISK              & \emph{k}-Means        & 0.7427 $\pm$ 0.1575  \\
FAST               & \emph{k}-Means        & 0.7376 $\pm$ 0.1602 \\
SURF               & \emph{k}-Means        & 0.7507 $\pm$ 0.1671  \vspace{0.05in} \\ \hline
BRISK              & FCM  & 0.7503 $\pm$ 0.1507  \\
FAST               & FCM  & 0.7379 $\pm$ 0.1601   \\
SURF               & FCM  & 0.7510 $\pm$ 0.1670
\end{tabular}
\end{table}

\section{Discussion and Conclusion}

In this study, a semi-automated localization method was proposed to mark a tumour with an ellipse based on a user single-click input, keypoint descriptors, and the SVM classification. The results demonstrated that the SURF features extended with the proposed weighted distances from the tumour's centre pixel (identified by the expert), along with the SVM classifier, achieved promising results. The original breast cancer ultrasound images exhibited very poor quality. Employing the central tumour point provided a higher correlation between the tumour area and the extracted features in noisy ultrasound images. Nonetheless, the segmentation of these images remains a challenging task as also reported by other researchers. For example, a recently published paper reported only between 50\% and 60\% accuracy in the segmentation of B-mode US images of breast~\cite{misc:Elawady16}.

In future work, a larger image dataset will be used for the training process. This will increase the chance of including examples of the images of different sizes and shapes, which will help to fully automate the approach.

\bibliographystyle{IEEEbib}
\bibliography{refs}

\begin{thebibliography}{10}

\bibitem{PM:Collins15}
F.~S. Collins and H.~Varmus,
\newblock ``A new initiative on precision medicine,''
\newblock {\em The New England Journal of Medicine}, vol. 372, no. 9, pp.
  793--795, 2015.

\bibitem{MI:Brindle08}
K.~Brindle,
\newblock ``{New approaches for imaging tumour responses to treatment},''
\newblock {\em Nature Reviews Cancer}, vol. 8, no. 2, pp. 94--107, 2008.

\bibitem{MI:Oelze16}
M.~L. Oelze and J.~Mamou,
\newblock ``Review of quantitative ultrasound: Envelope statistics and
  backscatter coefficient imaging and contributions to diagnostic ultrasound,''
\newblock {\em IEEE Trans. on Ultrasonics, Ferroelectrics, and Frequency
  Control}, vol. 63, no. 2, pp. 336--351, Feb. 2016.

\bibitem{Gangeh:TMI16}
M.~J. Gangeh, H.~Tadayyon, L.~Sannachi, A.~Sadeghi-Naini, W.~T. Tran, and G.~J.
  Czarnota,
\newblock ``Computer aided theragnosis using quantitative ultrasound
  spectroscopy and maximum mean discrepancy in locally advanced breast
  cancer,''
\newblock {\em IEEE Trans. on Medical Imaging}, vol. 35, no. 3, pp. 778--790,
  2016.

\bibitem{MI:Tadayyon15}
H.~Tadayyon, L.~Sannachi, M.~J. Gangeh, A.~Sadeghi-Naini, M.~Trudeau, and G.~J.
  Czarnota,
\newblock ``Early prediction of breast tumour response to chemotherapy using
  multiparametric quantitative ultrasound,''
\newblock in {\em 40$^{\textup{th}}$ International Symposium on Ultrasonic
  Imaging and Tissue Characterization (UITC)}, 2015.

\bibitem{Gangeh:SPIE16}
M.~J. Gangeh, A.~Raheem, H.~Tadayyon, S.~Liu, F.~Hadizad, and G.~J. Czarnota,
\newblock ``Breast tumour visualization using 3-{D} quantitative ultrasound
  methods,''
\newblock in {\em SPIE Medical Imaging}, 2016.

\bibitem{misc:Sridevi}
S.~Sridevi and M.~Sundaresan,
\newblock ``Survey of image segmentation algorithms on ultrasound medical
  images,''
\newblock in {\em International Conference on Pattern Recognition, Informatics
  and Mobile Engineering}, 2013, pp. 215--220.

\bibitem{misc:lowe04}
D.~G. Lowe,
\newblock ``Distinctive image features from scale-invariant keypoints,''
\newblock {\em International Journal of Computer Vision}, vol. 60, no. 2, pp.
  91--110, 2004.

\bibitem{misc:Rosten05}
E.~Rosten and T.~Drummond,
\newblock ``Fusing points and lines for high performance tracking,''
\newblock in {\em 10$^{\textup{th}}$ IEEE International Conference on Computer
  Vision (ICCV)}, Oct. 2005, vol.~2, pp. 1508--1515.

\bibitem{misc:Leutenegger11}
S.~Leutenegger, M.~Chli, and R.~Y. Siegwart,
\newblock ``Brisk: Binary robust invariant scalable keypoints,''
\newblock in {\em 13$^{\textup{th}}$ International Conference on Computer
  Vision (ICCV)}, Nov. 2011, pp. 2548--2555.

\bibitem{misc:Bay08}
H.~Bay, A.~Ess, T.~Tuytelaars, and L.~Van Gool,
\newblock ``Speeded-up robust features ({SURF}),''
\newblock {\em Computer Vision and Image Understanding}, vol. 110, no. 3, pp.
  346--359, 2008.

\bibitem{misc:Rosten10}
E.~Rosten, R.~Porter, and T.~Drummond,
\newblock ``Faster and better: A machine learning approach to corner
  detection,''
\newblock {\em IEEE Trans. on Pattern Analysis and Machine Intelligence}, vol.
  32, no. 1, pp. 105--119, 2010.

\bibitem{Tizhoosh1995}
H.~R. Tizhoosh and M.~Fochem,
\newblock ``Image enhancement with fuzzy histogram hyperbolization,''
\newblock {\em Proceedings of EUFIT}, pp. 1695--1698, 1995.

\bibitem{Labatut2015arxiv}
V.~Labatut and H.~Cherifi,
\newblock ``Accuracy measures for the comparison of classifiers,''
\newblock in {\em arXiv preprint}, 2012, vol. arXiv:1207.3790.

\bibitem{misc:Elawady16}
M.~Elawady, I.~Sadek, A.~E.~R. Shabayek, G.~Pons, and S.~Ganau,
\newblock ``Automatic nonlinear filtering and segmentation for breast
  ultrasound images,''
\newblock in {\em 13$^{\textup{th}}$ International Conference on Image Analysis
  and Recognition}, Berlin, Heidelberg, 2016, pp. 206--213, Springer-Verlag.

\end{thebibliography}
%
%\end{document}
%
%\end{thebibliography}

\end{document}